# AIRI: Predicting Retention Indices and their Uncertainties using Artificial Intelligence

Lewis Y. Geer[1*], Stephen E. Stein[1], William Gary Mallard[1], Douglas J. Slotta[1]

[1]*National Institute of Standards and Technology, 100 Bureau Dr., Gaithersburg, Maryland 20899, United States*

[*]Email: lewis.geer@nist.gov

## Abstract

The Kováts Retention index (RI) is a quantity measured using gas chromatography and commonly used in the identification of chemical structures. Creating libraries of observed RI values is a laborious task, so we explore the use of a deep neural network for predicting RI values from structure for standard semipolar columns. This network generated predictions with a mean absolute error of 15.1 and, in a quantification of the tail of the error distribution, a 95th percentile absolute error of 46.5. Because of the Artificial Intelligence Retention Indices (AIRI) network's accuracy, it was used to predict RI values for the NIST EI-MS spectral libraries. These RI values are used to improve chemical identification methods and the quality of the library.

Estimating uncertainty is an important practical need when using prediction models. To quantify the uncertainty of our network for each individual prediction, we used the outputs of an ensemble of 8 networks to calculate a predicted standard deviation for each RI value prediction. This predicted standard deviation was corrected to follow the error between observed and predicted RI values. The Z scores using these predicted standard deviations had a standard deviation of 1.52 and a 95th percentile absolute Z score corresponding to a mean RI value of 42.6.

## Introduction

The Kováts Retention index[1] (RI) is a widely used measure of gas chromatographic retention times relative to the retention times of *n*-paraffins of similar mass. RI values and the mass spectra of unknown chemicals are commonly measured in tandem to identify the chemicals through comparison to reference libraries containing both the mass spectra and RI values of known compounds. While we have maintained a multi-year program to measure a significant number of RI values matched to compounds for inclusion in the NIST/EPA/NIH mass spectral reference library[2–4], there remain a significant fraction of entries without measured values. In recent editions of this library, we have augmented these measured values with predictions from a group connectivity method[5] and more lately the deep neural network described in this paper to estimate RI values from molecular structure. Since these values are predictions, we report here newly developed methods to estimate the uncertainty of these predictions.

A retention index is determined by normalizing the retention time of a vaporized analyte passing through a column using selected reference compounds with retention times similar to the analyte. The Kováts index uses *n*-paraffins as these reference compounds. Modern columns are generally fused silica capillary columns coated with a stationary phase. This stationary phase consists of compounds that interact with the analyte through a variety of intermolecular forces, mechanisms of adsorption, and by steric constraints[6] that affect the adsorption of the analyte into the stationary phase. An increase in adsorption results in longer retention times. These RI values are highly reproducible and dependent on molecular structure, making them a useful quantity for structural identification by comparison to known standards.

Multiple approaches to estimating RI values have been described in the literature. Quantitative Structure Activity Relationship (QSAR) analyses have been used to predict RI values[6–12] with a large number of these analyses focused on specific classes of molecules[13]. Stein *et al.*[5] used group additivity to estimate the RI values of diverse compounds passed through polar and nonpolar columns with mean absolute prediction errors of 46 and 65, respectively. The use of deep learning methods for improving upon QSAR has been a topic of investigation since the start of the deep learning revolution[14] and has grown to encompass a large number of methods[15,16]. A number of these deep learning QSAR methods have been used to predict RI values[13,17–28] from molecular structure with varying degrees of success.

Since group additivity methods yield reasonable estimates of RI values, we set out to examine deep neural net architectures inspired by the group additivity approach from Stein *et al.*[5] The group additivity results imply that each layer of the architecture ought to have the ability to perceive for each atom in a structure not only the atoms bonding to it, but also the atoms that are more distant neighbors of the atom. This suggests that each atom should be characterized by not only its atomic number, but a variety of other structural concerns such as branching, nearest neighbor atoms, whether it is part of an aromatic system, and whether it is in a ring or not. Furthermore, the architecture ought to use this information to be able to select those atoms it believes are part of a group and ignore those that are not part of a group. This reasoning led us to exploring the use of Path-Augmented Graph Transformer Networks (PAGTN)[29], which can examine long range paths in a graph and uses an attention mechanism[30] to ignore or emphasize atoms in the paths. A hyperparameter search was performed to adjust the architecture of the PAGTN network to minimize the errors in RI prediction.

In the same sense that standard measurements become more useful when stated with an error, the practical use of RI predictions requires quantification of uncertainty. This need, however, is often left unexplored beyond aggregations performed on entire test sets. The most common aggregation is the Mean Absolute Error (MAE), which is the mean of the absolute difference between observed and predicted values. However, the MAE does not communicate the typically heavytailed distribution of RI predictions nor the variations due to individual chemical structures, which we quantify using the 95th percentile of the errors For example, a network that generates predictions with an MAE of 15 but with a 95th percentile absolute error of 70 can be less useful in terms of accuracy than a network that generates predictions with an MAE of 20 but a 95th percentile absolute error of 34. Furthermore, simple aggregations over the entire test set do not reveal uncertainties due to structural differences. To address this issue, we explore methods to yield a per structure uncertainty prediction that is less heavy-tailed.

# Methods

## Data

The 2023 NIST spectral reference libraries include measured standard semipolar RI values for 142,462 compounds. This data was randomly partitioned into training, validation and test sets. Compounds with RI values exceeding 6280 were omitted from these datasets as there are few RI values above this cutoff. Table 1 breaks down the number of compounds in the datasets. This breakdown includes whether the compounds were derivatized by trimethylsilane or not to allow separate examination of these groups since some RI prediction networks make poorer predictions for underivatized compounds.

**Table 1: Numbers of compounds in the datasets[a]**

|            | All     | TMS    | no TMS  |
|------------|---------|--------|---------|
| Train      | 132,440 | 24,461 | 107,979 |
| Validation | 2870    | 531    | 2339    |
| Test       | 7152    | 1389   | 5763    |

[a] "All" means all compounds, "TMS" means the compound is derivatized with trimethylsilane, and "no TMS" are compounds not derivatized with trimethylsilane.

## Standardization of compound structures

The structures of the compounds in the datasets were processed through a standardization method based on the RDKit[31] MolStandardize library to ensure the uniformity of the data input to the AIRI network. In the first step, the chemical structures were Kekulized, aromatized, conjugated and hybridized using standard rules. Valencies were not checked for allowed values since the V2000 version of the MOLfile format is largely limited to covalent bonds. Hydrogen atoms not explicitly necessary for the definition of the chemical structure were made implicit (that is, as an atom feature rather than a bond feature) and functional groups were corrected to standard forms. Finally, stereochemistry was recalculated. The precise rules for these computations can be found in the RDKit library, version 2023.03.2.

## PAGTN Architecture and Usage

The PAGTN architecture used in this study takes embedded atom and path features as input. The atom features are atomic number, formal charge, and the number of bonded atoms, which includes the number of hydrogens bonded implicitly and non-implicitly. The path features for each bond in the path include: bond type, bond conjugation, the shortest topological distance along the path, and whether the bond is in a ring. If a particular bond is in a ring, the features include the size of the ring and whether or not the ring is aromatic. The path features were truncated to a maximum allowed path length.

The architecture hyperparameters included the number of heads, the number of layers, the maximum path size, and the size of the hidden and query sublayers. Atoms did not attend to themselves and attention was global. Attention for a particular layer was only applied using the output from the previous layer. The output of the network was a summation of the final hidden layer. The RI values used for the output were divided by 10,000. 32-bit precision was used throughout the network.

For training, we investigated the use of the Adam or AdamW optimizer methods to see if either performed better. The training was performed with gradient clipping, a dropout of 0.2, and a learning rate of $5 \times 10^{-4}$. An MAE loss was used. The batch size was 50. The network contains approximately 2.6M parameters.

## Software Libraries

The evaluation methods and PATGN were coded in Python using the PyTorch[32], Lightning[33], Arrow, SciPy[34], Pandas[35], RDKit[31], and Numpy[36] libraries. Networks were trained on NVIDIA DGX-1 nodes. The network code and documentation can be found at https://github.com/usnistgov/masskit and https://github.com/usnistgov/masskit_ai, including a description of how to predict RI values from the command line. While use of the prediction network is more efficient using a GPU, our libraries can run on a CPU as detailed in the documentation.

# Results and Discussion

## Hyperparameter search

To maximize the ability of our network to predict RI values, we performed a search over a set of hyperparameters that determine the architecture of the network. The objective of this manually directed search was to minimize the MAE of predictions done on the validation set. This search was done over the following hyperparameters and ranges: network depths of 4 to 10, one and two attention heads, maximum path lengths of 2 to 6, and query and hidden layer sizes of 120 to 400 in steps of 40. The optimal hyperparameter settings were found to be a depth of 8, one attention head, a maximum path length of 5, and a query and hidden layer size of 280. These settings improved the MAE of the validation set by 6%.

Both Adam and AdamW optimizers were evaluated, with Adam providing marginally better results.

## Performance of the network

A standard method to improve the prediction accuracy of AI models is to average the predictions of multiple independently trained versions of models with the same architecture. For AIRI predictions, we created an ensemble of 8 networks, each independently trained using the same hyperparameters and the same training set randomized differently. In order to make a prediction, the same structure is fed into each network, and the resulting predicted retention index is averaged from the outputs of all of the models.

Using this approach, we generated predictions from the test set. The MAE of the test set predictions was 15.1 with a correlation coefficient of 0.9987. Since some RI prediction models perform differently on derivatized structures, we examined molecules derivatized with trimethylsilane (TMS). These had a MAE of 15.4, which was essentially the same as the MAE of underivatized molecules, 15.0. The MAE, the correlation coefficients, and the mean of the percentile confidence limit values are shown in Table 2. In comparison, the absolute difference between repeated measurements of RI values in our libraries has a median value of 3.8 and a 75th percentile value of 9.2.

**Table 2: Differences between predicted and observed RI values**[a]

|        | MAE  | 50% | 90%  | 95%  | 99%   | Correlation |
|--------|------|-----|------|------|-------|-------------|
| All    | 15.1 | 8.0 | 30.2 | 46.5 | 135.7 | 0.9987      |
| TMS    | 15.4 | 9.2 | 32.2 | 45.7 | 94.7  | 0.9987      |
| no TMS | 15.0 | 7.7 | 29.6 | 47.1 | 140.3 | 0.9987      |

[a] "MAE" is the mean absolute error. The percentages are the percentile absolute errors. "Correlation" is the correlation coefficient. "All" means all compounds, "TMS" means the compound is derivatized with trimethylsilane, and "no TMS" are compounds not derivatized with trimethylsilane.

---

It is important to note that the MAE, the median absolute error, and correlation coefficient can obscure the distribution of errors, particularly at the tails. Prediction models for RI can have heavy tails[5] so it is important to understand the entire distribution of errors as models with a small MAE value can still generate poor predictions for a large number of structures. To explore the distribution of errors at the tail, we calculated the 90th and 95th percentile absolute errors, that is, the error values where 90 or 95 percent of the absolute errors are less than or equal to the values. The absolute error is the absolute value of the observed value minus the predicted value. The 90th percentile absolute error was ±30.2 and the 95th percentile absolute error was ±46.5, indicating somewhat heavy-tailed but still useful predictions. The 99th percentile absolute error of ±135.7 indicates that there are significant outliers for a small percentage of the predictions, although some of these outliers may be due to errors in the data. A histogram of the test set errors and the percentiles of the absolute errors are shown in Figure 1.

**Figure 1**

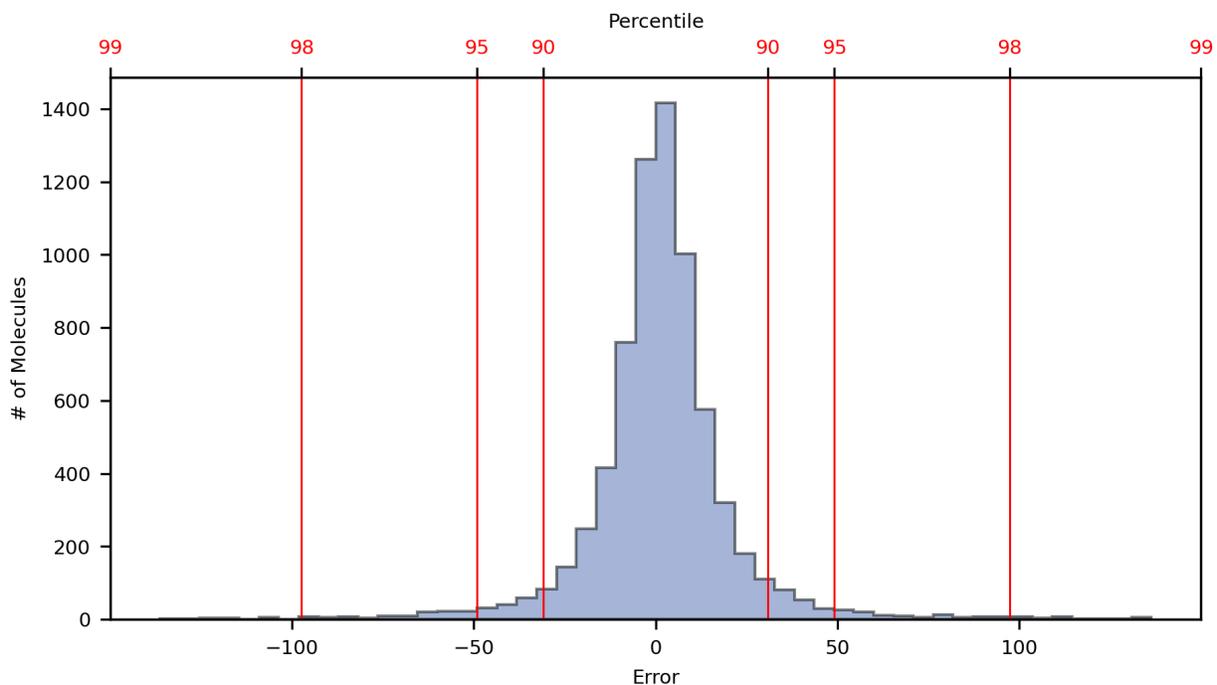

Figure 1. Histogram of the errors of the predicted RI values compared to the observed RI values in the test set. Shown in red are the percentiles of the absolute error.

## Estimating the uncertainty of the predictions

Our use of an ensemble of eight networks allows us to estimate the systematic error of the overall network itself, an approach being investigated in molecular property prediction[37,38]. This is done by calculating the predicted standard deviation of the multiple predictions given by our ensemble of networks for each predicted RI value. Examining the validation set, the correlation coefficient of the predicted standard deviation of the prediction to the prediction error was mildly correlated, with a value of 0.32. In Figure 2, we have plotted the observed and predicted RI values of a randomly selected subset of the test data, with the predicted standard deviation as error bars. These results indicate that the predicted standard deviation is a useful partial measure of the systematic error of the AIRI model.

**Figure 2**

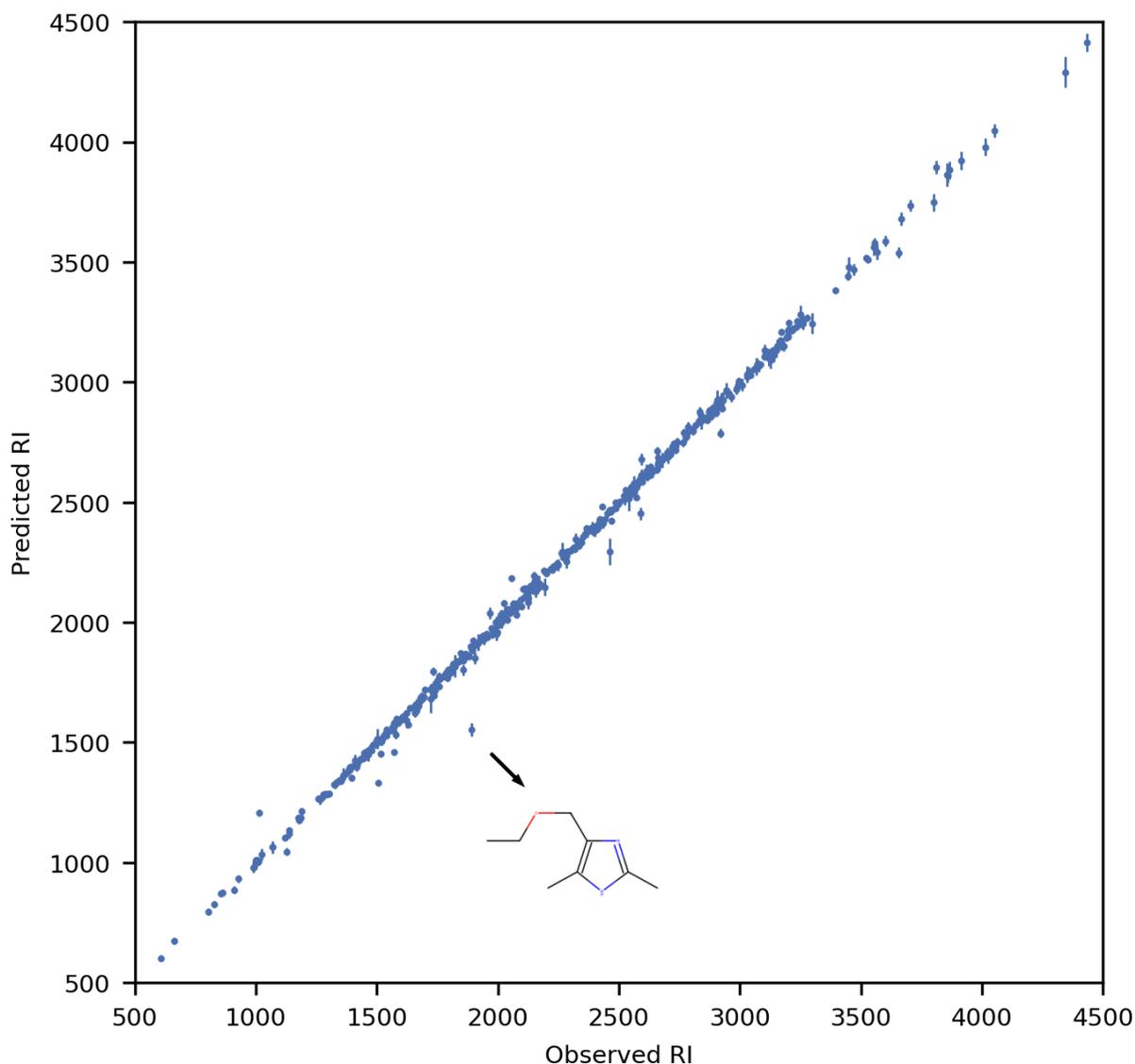

Figure 2. Plot of observed versus predicted RI values for 500 molecules randomly selected from the test set. Error bars are the corrected predicted standard deviations. For the sake of curiosity, the molecular structure of the furthest outlier is shown at the bottom of the figure.

In an attempt to understand the sources of systematic error, we examined the correlation between the error in the validation set and the 2D molecular descriptors found in the RDKit cheminformatics library. Only weak correlations, with correlation coefficients of order 0.2, were found. These correlations were to nitrogen related descriptors, such as tertiary amines, and BertzCT, a topological measure of molecular complexity. Performing linear regression of these descriptors in order to improve the RI predictions did not yield significant results. However, these weak correlations may be useful in improving or defining the architecture of future models.

## Correcting the predicted standard deviation to improve the tails of the error distribution

Since RI predictions tend to be heavy-tailed, we performed an analysis to reduce the size of the tails of the error distribution when expressed as a Z score, which takes into account per structure variation in the error. This is pragmatically important as an excess of outliers makes these predictions less useful -- even if most predictions have minimal error, the predictions with greater error will often determine whether results are to be trusted for a particular application. Z scores have the form $Z = (x-\mu)/\sigma$ where $x$ is the observed RI value, $\mu$ is the mean predicted RI value of the network ensemble, and $\sigma$ is the predicted standard deviation. Since the Z score incorporates the predicted standard deviation, it allows for an error that can be expressed per structure. A histogram of the Z scores for the test set is shown in dotted outline in Figure 3. While the histogram does appear similar to a standardized normal distribution with a standard distribution of 2.39, it is heavy-tailed.

**Figure 3**

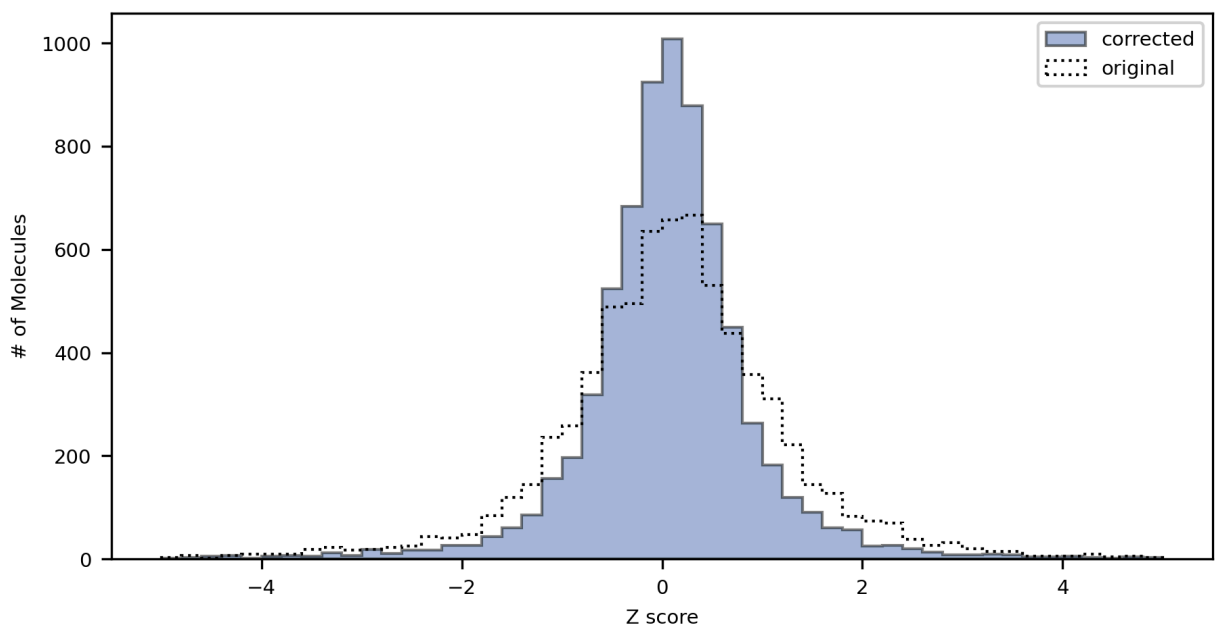

Figure 3. Histogram of the Z scores of the predicted RI values compared to the observed RI values in the test set. For the original histogram shown in dotted outline, the standard deviation used to calculate the Z score is the predicted standard deviation. For the corrected histogram shown in blue, the standard deviation used to calculate the Z score is the corrected predicted standard deviation.

Examination of the predicted standard deviation in comparison to the actual absolute error between observed and predicted RI values shows that for small and large values of the predicted standard deviation it is consistently smaller than the actual absolute error. To quantify and correct for this issue, we examined the predicted standard deviation by dividing the range of the predicted standard deviation into a set of bins applied over the training set, where the bin size $b$ was varied to find the optimal Z score correction. For each bin, we calculated the ratio of the $p$th percentile of the absolute error (observed RI values minus the predicted RI values) divided by the $p$th percentile of the predicted standard deviation. $p$ was also varied to find the optimal Z score correction. To find the ideal values of $p$ and $b$, we allowed $p$ to range from 50% to 99% in steps of 1% and $b$ to range from 3 to 24. The resulting set of absolute error to predicted standard deviation ratios was then used to correct each predicted standard deviation in the validation set by multiplying each predicted standard deviation by the value of the ratio for the corresponding bin. The corrected predicted standard deviation then was used to calculate a corrected Z score. We computed the standard deviation of the corrected Z score as well as the 95th percentile corrected Z scores, which were then used to calculate a corresponding mean RI value. We selected a value of $p$ and $b$ that minimized both the standard deviation of the corrected Z score and mean RI value of the 95th percentile corrected Z scores. For the validation set, these values were $p=78\%$ and $b=2$, with the corresponding correction ratio plotted in Figure 4. Note that the predicted standard deviation is smaller than the absolute error for low and high values.

The correction was then applied to the test set and the resulting corrected Z scores are shown as the blue histogram in Figure 3, showing that the resulting Z score distribution is more normally distributed with a standard deviation of 1.51 and an 95th percentile absolute Z score of 2.046. This 95th percentile absolute Z score corresponds to a mean RI value of 42.6, which is improved from the uncorrected value.

**Figure 4**

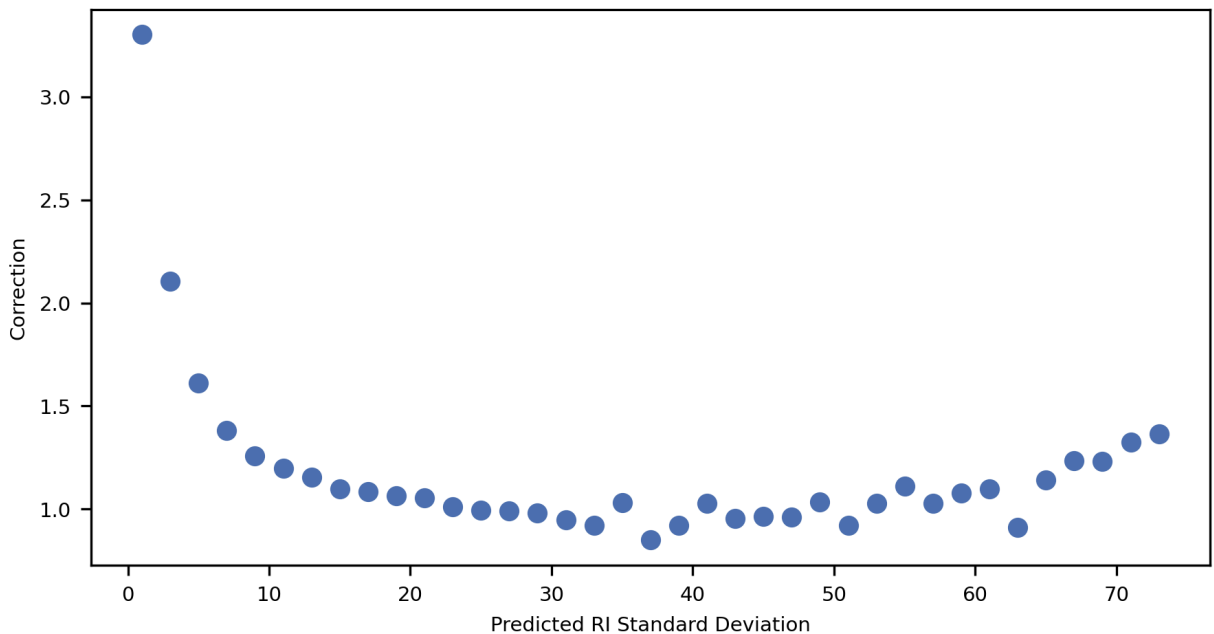

Figure 4. Ratio of the 78th percentile absolute error divided by the 78th percentile predicted RI standard deviation. The ratio is calculated for a range of predicted RI standard deviations in bins of size 2 and used to correct the predicted standard deviation.

## Conclusions

To augment existing libraries of RI values and for quality control, we explored the use of a PATGN deep neural network called AIRI for predicting RI values, performing a comprehensive hyperparameter search to optimally structure the network. The resulting network was ensembled and generated predictions with a MAE of 15.1 and, more importantly as RI prediction models tend to have long tails, a 95th percentile absolute error of 46.5. While a comparison to other models is not in the scope of this paper, Vrzal et al.[22] reported an MAE of 28.4 with a 95th percentile absolute error of 67.6 for DeepRel and an MAE of 46.9 and 95th percentile absolute error of 136.2 for the Matyushin et al.[17] model. Anjum et al.[18] reported an MAE of 16.57 for derivatized compounds and an MAE of 29.55 for underivatized compounds for the RIpred model. In the same study they evaluated the Qu et al.[25] model to have an MAE of 16.84 for underivatized compounds and 22.56 for derivatized compounds. Note that the Qu et al. model has never been used for NIST spectral libraries as asserted by Anjum et al. They did not report a long tail statistic, such as 95th percentile, for either model.

Effective use of retention index values requires a knowledge of their uncertainty. For each predicted RI value we used the outputs of the network ensemble to calculate a predicted standard deviation. This predicted standard deviation was then corrected to more closely match the error of the prediction by the use of a correction coefficient that varies with the value of the predicted standard deviation. To measure the quality of these corrected predicted standard deviations, the Z scores from these standard deviations improved from a standard deviation of 2.39 to 1.52 with a mean of the 95% percentile absolute Z score that corresponds to a mean RI value of 42.6.

The predictions from the AIRI model have found multiple uses in building spectral libraries and in searching them. For quality assurance, AIRI predictions are used to check for errors in measured RI values and in RI values from the literature, as well as deciding which derivatized structure of an analyte was injected into a mass spectrometer or whether the original compound decomposed prior to detection. AIRI predictions can improve the process of chemical identification, including improving the order of hitlists from spectral library searches. Because the corrected prediction error is within a few multiples of experimental error for almost all predicted RI values, histlists can be winnowed by omitting or downweighting spectral matches to molecular spectra whose corresponding RI measurements are poorly matched to predicted or observed RI values in the spectral library. Because of all of these advantages, we have used variations of this PAGTN network to predict RI values for the widely used 2020 and 2023 NIST EI-MS spectral libraries.

A command line program that predicts AIRI values from MOLfiles and SMILES has been publicly released on GitHub, as described in the Methods. Future work on the AIRI model includes the improvement of long tail predictions and the prediction of RI values for other standard column types.

## Associated Content

### Data Availability Statement

The source code for this study can be found in two GitHub repositories: https://github.com/usnistgov/masskit and https://github.com/usnistgov/masskit_ai. The public distribution of the data for this study is described at https://chemdata.nist.gov/dokuwiki/doku.php?id=chemdata:ridatabase.

## Author Information

### Corresponding Author

**Lewis Y. Geer** --- *National Institute of Standards and Technology, 100 Bureau Dr., Gaithersburg, Maryland 20899, United States*; Phone: +1 301 975 6820; https://orcid.org/0000-0001-8189-3821; Email: lewis.geer@nist.gov

### Authors

**William Gary Mallard** --- *National Institute of Standards and Technology, 100 Bureau Dr., Gaithersburg, Maryland 20899, United States*; https://orcid.org/0000-0003-2158-5098

**Douglas J. Slotta** --- *National Institute of Standards and Technology, 100 Bureau Dr., Gaithersburg, Maryland 20899, United States*; https://orcid.org/0009-0000-9085-4891

**Stephen E. Stein** --- *National Institute of Standards and Technology, 100 Bureau Dr., Gaithersburg, Maryland 20899, United States*; https://orcid.org/0000-0001-9384-3450

### Notes

The authors declare no competing financial interests. All commercial instruments, software, and materials used in the study are for experimental purposes only. Such identification does not intend recommendation or endorsement by the National Institute of Standards and Technology, nor does it intend that the materials, software, or instruments used are necessarily the best available for the purpose. This work was supported by the National Institute of Standards and Technology.

## Acknowledgements

We thank Dr. William E. Wallace for many useful comments. This paper is dedicated to the memory of Ira W. Geer.

## References

(1) Kováts, E. Gas‐chromatographische Charakterisierung Organischer Verbindungen. Teil 1: Retentionsindices Aliphatischer Halogenide, Alkohole, Aldehyde Und Ketone. *Helv. Chim. Acta* **1958**, *41*, 1915–1932. https://doi.org/10.1002/hlca.19580410703.

(2) Ausloos, P.; Clifton, C. L.; Lias, S. G.; Mikaya, A. I.; Stein, S. E.; Tchekhovskoi, D. V.; Sparkman, O. D.; Zaikin, V.; Zhu, D. The Critical Evaluation of a Comprehensive Mass Spectral Library. *J. Am. Soc. Mass Spectrom.* **1999**, *10*, 287–299. https://doi.org/10.1016/S1044-0305(98)00159-7.

(3) Babushok, V. I.; Linstrom, P. J.; Reed, J. J.; Zenkevich, I. G.; Brown, R. L.; Mallard, W. G.; Stein, S. E. Development of a Database of Gas Chromatographic Retention Properties of Organic Compounds. *J. Chromatogr. A* **2007**, *1157*, 414–421. https://doi.org/10.1016/j.chroma.2007.05.044.

(4) NIST Mass Spectrometry Data Center. NIST EI-MS Spectral Library, 2023. https://chemdata.nist.gov/dokuwiki/doku.php?id=chemdata:ridatabase.

(5) Stein, S. E.; Babushok, V. I.; Brown, R. L.; Linstrom, P. J. Estimation of Kováts Retention Indices Using Group Contributions. *J. Chem. Inf. Model.* **2007**, *47*, 975–980. https://doi.org/10.1021/ci600548y.

(6) Peng, C. T. Prediction of Retention Indices: V. Influence of Electronic Effects and Column Polarity on Retention Index. *J. Chromatogr. A* **2000**, *903*, 117–143. https://doi.org/10.1016/S0021-9673(00)00901-8.

(7) Peng, C. T.; Yang, Z. C.; Ding, S. F. Prediction of Retention Indexes: II. Structure-Retention Index Relationship on Polar Columns. *J. Chromatogr. A* **1991**, *586*, 85–112. https://doi.org/10.1016/0021-9673(91)80028-F.

(8) Garkani-Nejad, Z.; Karlovits, M.; Demuth, W.; Stimpfl, T.; Vycudilik, W.; Jalali-Heravi, M.; Varmuza, K. Prediction of Gas Chromatographic Retention Indices of a Diverse Set of Toxicologically Relevant Compounds. *J. Chromatogr. A* **2004**, *1028*, 287–295. https://doi.org/10.1016/j.chroma.2003.12.003.

(9) Budahegyi, M. V.; Lombosi, E. R.; Lombosi, T. S.; Mészáros, S. Y.; Nyiredy, Sz.; Tarján, G.; Timár, I.; Takács, J. M. Twenty-Fifth Anniversary of the Retention Index System in Gas—Liquid Chromatography. *J. Chromatogr. A* **1983**, *271*, 213–307. https://doi.org/10.1016/S0021-9673(00)80220-4.

(10) Randić, M.; Basak, S. C.; Pompe, M.; Novič, M. Prediction of Gas Chromatographic Retention Indices Using Variable Connectivity Index. *Acta Chim Slov.* **2001**, *48*, 169–180.

(11) Price, G. J.; Dent, M. R. Prediction of Retention in Gas-Liquid Chromatography Using the UNIFAC Group Contribution Method: II. Polymer Stationary Phases. *J. Chromatogr. A* **1991**, *585*, 83–92. https://doi.org/10.1016/0021-9673(91)85060-S.

(12) Peng, C. T.; Ding, S. F.; Hua, R. L.; Yang, Z. C. Prediction of Retention Indexes: I. Structure—Retention Index Relationship on Apolar Columns. *J. Chromatogr. A* **1988**, *436*, 137–172. https://doi.org/10.1016/S0021-9673(00)94575-8.

(13) Goel, P.; Bapat, S.; Vyas, R.; Tambe, A.; Tambe, S. S. Genetic Programming Based Quantitative Structure–Retention Relationships for the Prediction of Kovats Retention Indices. *J. Chromatogr. A* **2015**, *1420*, 98–109. https://doi.org/10.1016/j.chroma.2015.09.086.

(14) Dahl, G. E.; Jaitly, N.; Salakhutdinov, R. Multi-Task Neural Networks for QSAR

Predictions. arXiv June 4, 2014. https://doi.org/10.48550/arXiv.1406.1231.
(15) Li, Z.; Jiang, M.; Wang, S.; Zhang, S. Deep Learning Methods for Molecular Representation and Property Prediction. *Drug Discov. Today* **2022**, *27*, 103373. https://doi.org/10.1016/j.drudis.2022.103373.
(16) Ramsundar, B.; Eastman, P.; Walters, P.; Pande, V. *Deep Learning for the Life Sciences*; O'Reilly Media, Inc., 2019.
(17) Matyushin, D. D.; Sholokhova, A. Yu.; Buryak, A. K. A Deep Convolutional Neural Network for the Estimation of Gas Chromatographic Retention Indices. *J. Chromatogr. A* **2019**, *1607*, 460395. https://doi.org/10.1016/j.chroma.2019.460395.
(18) Anjum, A.; Liigand, J.; Milford, R.; Gautam, V.; Wishart, D. S. Accurate Prediction of Isothermal Gas Chromatographic Kováts Retention Indices. *J. Chromatogr. A* **2023**, *1705*, 464176. https://doi.org/10.1016/j.chroma.2023.464176.
(19) Kireev, A.; Osipenko, S.; Mallard, G.; Nikolaev, E.; Kostyukevich, Y. Comparative Prediction of Gas Chromatographic Retention Indices for GC/MS Identification of Chemicals Related to Chemical Weapons Convention by Incremental and Machine Learning Methods. *Separations* **2022**, *9*, 265. https://doi.org/10.3390/separations9100265.
(20) D'Archivio, A. A.; Maggi, M. A.; Ruggieri, F. Cross-Column Prediction of Gas-Chromatographic Retention Indices of Saturated Esters. *J. Chromatogr. A* **2014**, *1355*, 269–277. https://doi.org/10.1016/j.chroma.2014.06.002.
(21) Matyushin, D. D.; Sholokhova, A. Y.; Buryak, A. K. Deep Learning Based Prediction of Gas Chromatographic Retention Indices for a Wide Variety of Polar and Mid-Polar Liquid Stationary Phases. *Int. J. Mol. Sci.* **2021**, *22*, 9194. https://doi.org/10.3390/ijms22179194.
(22) Vrzal, T.; Malečková, M.; Olšovská, J. DeepReI: Deep Learning-Based Gas Chromatographic Retention Index Predictor. *Anal. Chim. Acta* **2021**, *1147*, 64–71. https://doi.org/10.1016/j.aca.2020.12.043.
(23) Matyushin, D. D.; Buryak, A. K. Gas Chromatographic Retention Index Prediction Using Multimodal Machine Learning. *IEEE Access* **2020**, *8*, 223140–223155. https://doi.org/10.1109/ACCESS.2020.3045047.
(24) de Cripan, S. M.; Cereto-Massagué, A.; Herrero, P.; Barcaru, A.; Canela, N.; Domingo-Almenara, X. Machine Learning-Based Retention Time Prediction of Trimethylsilyl Derivatives of Metabolites. *Biomedicines* **2022**, *10*, 879. https://doi.org/10.3390/biomedicines10040879.
(25) Qu, C.; Schneider, B. I.; Kearsley, A. J.; Keyrouz, W.; Allison, T. C. Predicting Kováts Retention Indices Using Graph Neural Networks. *J. Chromatogr. A* **2021**, *1646*, 462100. https://doi.org/10.1016/j.chroma.2021.462100.
(26) Dossin, E.; Martin, E.; Diana, P.; Castellon, A.; Monge, A.; Pospisil, P.; Bentley, M.; Guy, P. A. Prediction Models of Retention Indices for Increased Confidence in Structural Elucidation during Complex Matrix Analysis: Application to Gas Chromatography Coupled with High-Resolution Mass Spectrometry. *Anal. Chem.* **2016**, *88*, 7539–7547. https://doi.org/10.1021/acs.analchem.6b00868.
(27) Yan, J.; Huang, J.-H.; He, M.; Lu, H.-B.; Yang, R.; Kong, B.; Xu, Q.-S.; Liang, Y.-Z. Prediction of Retention Indices for Frequently Reported Compounds of Plant Essential Oils Using Multiple Linear Regression, Partial Least Squares, and Support Vector Machine. *J. Sep. Sci.* **2013**, *36*, 2464–2471.

(28) Chen, H.-F. Quantitative Predictions of Gas Chromatography Retention Indexes with Support Vector Machines, Radial Basis Neural Networks and Multiple Linear Regression. *Anal. Chim. Acta* **2008**, *609*, 24–36. https://doi.org/10.1016/j.aca.2008.01.003.
(29) Chen, B.; Barzilay, R.; Jaakkola, T. Path-Augmented Graph Transformer Network. arXiv May 29, 2019. https://doi.org/10.48550/arXiv.1905.12712.
(30) Vaswani, A.; Shazeer, N.; Parmar, N.; Uszkoreit, J.; Jones, L.; Gomez, A. N.; Kaiser, L.; Polosukhin, I. Attention Is All You Need. **2017**. https://doi.org/10.48550/ARXIV.1706.03762.
(31) RDKit: Open-Source Cheminformatics, 2023. https://www.rdkit.org.
(32) Paszke, A.; Gross, S.; Massa, F.; Lerer, A.; Bradbury, J.; Chanan, G.; Killeen, T.; Lin, Z.; Gimelshein, N.; Antiga, L.; Desmaison, A.; Kopf, A.; Yang, E.; DeVito, Z.; Raison, M.; Tejani, A.; Chilamkurthy, S.; Steiner, B.; Fang, L.; Bai, J.; Chintala, S. PyTorch: An Imperative Style, High-Performance Deep Learning Library. **2019**. https://doi.org/10.48550/arXiv.1912.01703.
(33) Falcon, W.; Borovec, J.; Wälchli, A.; Eggert, N.; Schock, J.; Jordan, J.; Skafte, N.; Ir1dXD; Bereznyuk, V.; Harris, E.; Tullie Murrell; Yu, P.; Præsius, S.; Addair, T.; Zhong, J.; Lipin, D.; Uchida, S.; Shreyas Bapat; Schröter, H.; Dayma, B.; Karnachev, A.; Akshay Kulkarni; Shunta Komatsu; Martin.B; Jean-Baptiste SCHIRATTI; Mary, H.; Byrne, D.; Cristobal Eyzaguirre; Cinjon; Bakhtin, A. PyTorchLightning/Pytorch-Lightning: 0.7.6 Release, 2020. https://doi.org/10.5281/ZENODO.3828935.
(34) Virtanen, P.; Gommers, R.; Oliphant, T. E.; Haberland, M.; Reddy, T.; Cournapeau, D.; Burovski, E.; Peterson, P.; Weckesser, W.; Bright, J.; van der Walt, S. J.; Brett, M.; Wilson, J.; Millman, K. J.; Mayorov, N.; Nelson, A. R. J.; Jones, E.; Kern, R.; Larson, E.; Carey, C. J.; Polat, İ.; Feng, Y.; Moore, E. W.; VanderPlas, J.; Laxalde, D.; Perktold, J.; Cimrman, R.; Henriksen, I.; Quintero, E. A.; Harris, C. R.; Archibald, A. M.; Ribeiro, A. H.; Pedregosa, F.; van Mulbregt, P. SciPy 1.0: Fundamental Algorithms for Scientific Computing in Python. *Nat. Methods* **2020**, *17*, 261–272. https://doi.org/10.1038/s41592-019-0686-2.
(35) McKinney, W. Data Structures for Statistical Computing in Python; Austin, Texas, 2010; pp 56–61. https://doi.org/10.25080/Majora-92bf1922-00a.
(36) Harris, C. R.; Millman, K. J.; van der Walt, S. J.; Gommers, R.; Virtanen, P.; Cournapeau, D.; Wieser, E.; Taylor, J.; Berg, S.; Smith, N. J.; Kern, R.; Picus, M.; Hoyer, S.; van Kerkwijk, M. H.; Brett, M.; Haldane, A.; del Río, J. F.; Wiebe, M.; Peterson, P.; Gérard-Marchant, P.; Sheppard, K.; Reddy, T.; Weckesser, W.; Abbasi, H.; Gohlke, C.; Oliphant, T. E. Array Programming with NumPy. *Nature* **2020**, *585*, 357–362. https://doi.org/10.1038/s41586-020-2649-2.
(37) Busk, J.; Jørgensen, P. B.; Bhowmik, A.; Schmidt, M. N.; Winther, O.; Vegge, T. Calibrated Uncertainty for Molecular Property Prediction Using Ensembles of Message Passing Neural Networks. *Mach. Learn. Sci. Technol.* **2021**, *3*, 015012. https://doi.org/10.1088/2632-2153/ac3eb3.
(38) Hirschfeld, L.; Swanson, K.; Yang, K.; Barzilay, R.; Coley, C. W. Uncertainty Quantification Using Neural Networks for Molecular Property Prediction. *J. Chem. Inf. Model.* **2020**, *60*, 3770–3780. https://doi.org/10.1021/acs.jcim.0c00502.

For Table of Contents Use Only

**AIRI: Predicting Retention Indices and their Uncertainties using Artificial Intelligence**

Lewis Y. Geer[*], Stephen E. Stein, W. Gary Mallard, Douglas J. Slotta

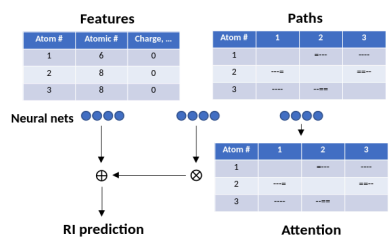